\documentclass[10pt,twocolumn,letterpaper]{article}

\usepackage[pagenumbers]{cvpr} % To force page numbers, e.g. for an arXiv version
\usepackage{amsmath,algpseudocode}

\usepackage{blindtext}
\usepackage{cuted}
\usepackage{placeins} 
\usepackage{booktabs}
\usepackage{multirow}
\usepackage[linesnumbered,ruled,vlined]{algorithm2e}
\DontPrintSemicolon
\SetKwComment{Comment}{\color{green!50!black}// }{}

\newcommand{\var}{\texttt}
\newcommand{\FuncCall}[2]{\texttt{\bfseries #1(#2)}}
\SetKwProg{Function}{function}{}{}

\definecolor{cvprblue}{rgb}{0.21,0.49,0.74}
\usepackage[pagebackref,breaklinks,colorlinks,citecolor=cvprblue]{hyperref}
\usepackage{animate}

\newcommand{\ourmodel}[1]{\textit{Pix2Gif}}

\def\paperID{10374} % *** Enter the Paper ID here
\def\confName{CVPR}
\def\confYear{2024}

\title{Pix2Gif: Motion-Guided Diffusion for GIF Generation}

\author{Hitesh Kandala\\
Microsoft Research, India\\
{\tt\small t-hkandala@microsoft.com}
\and
Jianfeng Gao\\
Microsoft Research, Redmond\\
{\tt\small jfgao@microsoft.com}
\and
Jianwei Yang\\
Microsoft Research, Redmond\\
{\tt\small jianwyan@microsoft.com}
}

\begin{document}
\maketitle
\begin{strip}
    \centering\noindent  
    \includegraphics[width=\textwidth]{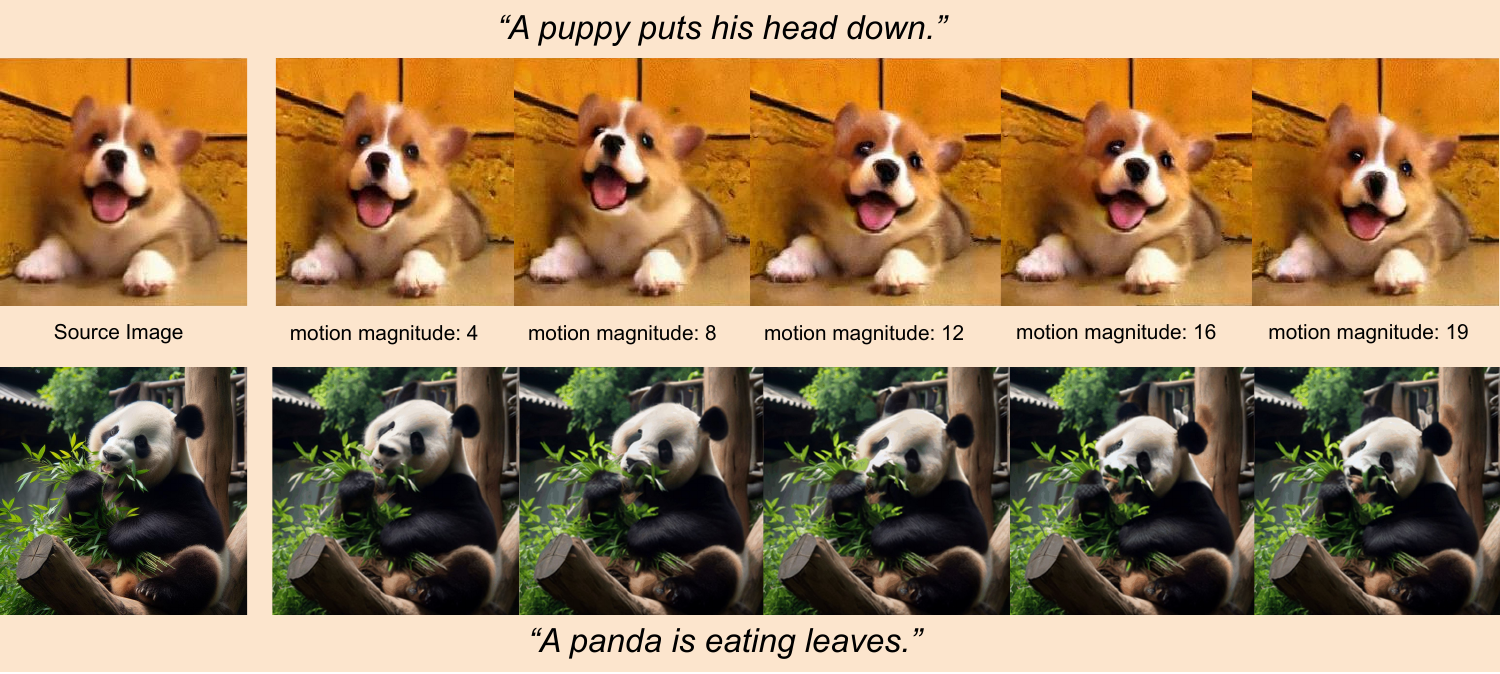}       
    \captionof{figure}{Our model creates distinct frames based on the provided source image and caption, adjusting according to different levels of motion magnitude (optical flow magnitude) specified in the input conditions. It stands well for both high spatial quality and temporal consistency.}
\label{fig:teaser}
\end{strip}
% \begin{figure*}
%   \centering
%    \includegraphics[width=0.8\linewidth]{images/teaser.png}

%    \caption{Example of caption.
%    It is set in Roman so that mathematics (always set in Roman: $B \sin A = A \sin B$) may be included without an ugly clash.}
%    \label{fig:dataset}
% \end{figure*}

\begin{abstract}
We present \textit{Pix2Gif}, a motion-guided diffusion model for image-to-GIF (video) generation. We tackle this problem differently by formulating the task as an image translation problem steered by text and motion magnitude prompts, as shown in Fig.~\ref{fig:teaser}. To ensure that the model adheres to motion guidance, we propose a new motion-guided warping module to spatially transform the features of the source image conditioned on the two types of prompts. Furthermore, we introduce a perceptual loss to ensure the transformed feature map remains within the same space as the target image, ensuring content consistency and coherence. In preparation for the model training, we meticulously curated data by extracting coherent image frames from the TGIF video-caption dataset, which provides rich information about the temporal changes of subjects. After pretraining, we apply our model in a zero-shot manner to a number of video datasets. Extensive qualitative and quantitative experiments demonstrate the effectiveness of our model -- it not only captures the semantic prompt from text but also the spatial ones from motion guidance. We train all our models using a single node of 16$\times$V100 GPUs. Code, dataset and models are made public at: \url{https://hiteshk03.github.io/Pix2Gif/}.
\end{abstract}    
\section{Introduction}
\label{sec:intro}

Visual content generation has been significantly advanced by the huge progress of diffusion models~\cite{lu2022dpmsolver,song2022denoising,ho2020denoising,ho2022classifier}. Recently, the development of latent diffusion models (LDMs) ~\cite{rombach2022highresolution} has led us to a new quality level of generated images. It has inspired a lot of works for customized and controllable image generation~\cite{zhang2023adding,li2023gligen,yang2022reco,ruiz2023dreambooth}, and fine-grained image editing~\cite{brooks2023instructpix2pix,hertz2022prompttoprompt,meng2022sdedit,kawar2023imagic}.

In this work, we focus on converting a single image to an animated Graphics Interchange Format (GIF), which is valuable for design yet under-explored. Despite the absence of image-to-GIF generation models, diffusion-based video generation has emerged as a hot topic recently. Compared with text-to-image generation, however, text-to-video generation requires not only high quality for individual frames but also visual consistency and temporal coherence across frames. To achieve this goal, existing works expand the LDMs to video diffusion models (VDMs) by either inflating the 2D CNNs in LDMs to 3D ones~\cite{ho2022video} or introducing an additional temporal attention layer to bridge the diffusion for each frame~\cite{ge2023preserve,he2023latent,blattmann2023align,singer2022makeavideo,wang2023lavie}. In addition to text prompts, a few recent works also explored the way of using images or other prompts to make the video generation model more customizable and controllable~\cite{esser2023structure,ni2023conditional}. However, due to the high cost of VDMs to generate a sequence of video frames in one run, most (if not all) of these works require a compromise of reducing the resolution of generated frames ($64\times 64$ typically), and the usage of extra super-resolution diffusion models for upscaling~\cite{li2022srdiff,saharia2022photorealistic}. Moreover, since these methods use the temporal attention layers to model the cross-frame dependency implicitly, it is quite hard for them to preserve good controllability of the frame-to-frame temporal dynamics in a fine-grained manner.

Given that animated GIF usually contains less number of frames and requires more specializations, we take a different strategy and formulate the image-to-GIF generation as an image translation process. To decouple the generation of visual contents and temporal dynamics, we further introduce a motion flow magnitude as extra guidance in addition to image and text prompts. Unlike the aforementioned works, our model takes one or more history frames as the condition and produces only one future frame at once. This brings some unique advantages: $(i)$ \emph{simplicity} - our model can be purely built on top of LDMs and trained end-to-end with high resolution, without any cascaded diffusion processes for upscaling. $(ii)$ \emph{controllability} - we could inject detailed and different text and motion prompts at each time step for generating a frame, which gains much better controllability of the model. Our work is inspired by a line of canonical works for future frame prediction~\cite{oliu2018folded,shi2015convolutional,villegas2018decomposing}. However, due to the lack of a powerful image-generation engine, these works fail to produce high-quality results and can only be applied to specific video domains~\cite{lee2019mutual,girdhar2020cater,soomro2012ucf101}. Moreover, they cannot support other types of prompts or conditions than the history frames. To address this problem, we exploit a modern diffusion-based pipeline. More specifically, we follow text-conditioned image editing approaches (\textit{e.g.}, InstructPix2Pix~\cite{brooks2023instructpix2pix}), and propose a new temporal image editing to produce future frames given history frames. To train the model, we curate a new training dataset based on TGIF~\cite{li2016tgif} by extracting frames and calculating the magnitude of optical flow between them. We then selected an appropriate range of the optical flow magnitude and sampled frame pairs from each GIF in a manner that ensures diversity. In the end, we train our diffusion model called \textit{Pix2Gif}, which can generate high-quality animated GIF consisting of multiple frames, given a single image and text and motion magnitude prompts. In summary, our main contributions are: 

% \begin{itemize}
%     \item start into
%     % \item We explore a different but intuitive approach of using just the source image to generate all the frames without complicating either the model or training paradigm
%     % \item Given an image and a text i.e. a caption to describe what the video should be about, our model generates next frames based on the amount of motion you need by inputting the magnitude of optical flow.
%     \item This is different than doing it auto-regressively where the major condition is on the earlier generated image.
%     \item We leverage a text-video pair dataset to train our model in the text-image training paradigm.
%     \item We use a warping module which captures and translates the source image to the target image based entirely on the caption and interval embedding. This warped source image provides a good prior to what the target image should look like.
% \end{itemize}
\begin{itemize}
    \item We are the first to explore an image-to-image translation formula for generating animated GIFs from an image, guided by a text prompt and motion magnitude.
    \item We propose a flow-based warping module with a perceptual loss in the diffusion process that takes motion magnitude as input and controls the temporal dynamics and consistency between future frames and the initial ones.
    \item We curate a new dataset, comprised of 78,692 short GIF clips for training, and 10,546 for evaluation. The new dataset covers a variety of visual domains.
    \item Quantitative and qualitative results demonstrate the effectiveness of our proposed method for generating visually consistent coherent GIFs from a single image, and it can be generalized to a wide range of visual domains.
\end{itemize}

\section{Prior Work}
\label{sec:prior_work}

Image and video generation has been a long-standing problem in the community. It can be tackled by different approaches, which can be categorized into four groups: generative adversarial networks (GANs)~\cite{goodfellow2020generative,radford2015unsupervised,kang2023scaling,karras2019style}, transformer-based autoregressive decoding~\cite{oord2016conditional,oord2016pixel,weissenborn2020scaling,ding2021cogview,yu2022scaling,esser2021taming}, masked image modeling~\cite{chang2023muse,chang2022maskgit,villegas2022phenaki,yu2023magvit}. Most of the recent works exploited diffusion models for image generation given their high-quality outputs and huge open-source supports~\cite{rombach2022highresolution,saharia2022photorealistic}. Recently, a number of works have extended the text-to-image generation model into image translation or editing models~\cite{brooks2023instructpix2pix,kawar2023imagic,zhang2023adding,meng2022sdedit,hertz2022prompttoprompt} or video generation models~\cite{blattmann2023align,ho2022video,singer2022makeavideo,voleti2022mcvd,ge2023preserve}. Below we provide a brief overview of the related diffusion-based image and video generation methods.

% \paragraph{Text-to-Image Generation.} 

\paragraph{Image-to-Image Translation.} Diffusion-based image-to-image generation has drawn increasing attention. Different from text-to-image generation, it takes an image as input and edits its contents following the text instructions while keeping the irrelated parts unchanged. SDEdit~\cite{meng2022sdedit} and ILVR~\cite{choi2021ilvr} are two pioneering works that impose reference image conditions to an existing latent diffusion model for controllable image generation. Later on, to conduct local edits, the authors in~\cite{Avrahami_2023} proposed blended latent diffusion to steer the diffusion process with a user-specified mask, where the pixels out of the mask remain the same as the input image while the region inside is edited following the textual description. Instead of manipulating the image space, Prompt2Prompt~\cite{hertz2022prompttoprompt} proposed to edit the image by manipulating the textual context (\textit{e.g.}, swapping or adding words.) to which the latent diffusion model cross-attends.  However, this method requires forwarding a text-to-image generation process to obtain the cross-attention maps, and thus cannot be applied to real images. Imagic~\cite{kawar2023imagic} proposed to blend the embeddings of a real image with the textual context embedding so that the generated image obeys both the image and text conditions.

All the aforementioned works leverage a frozen latent diffusion model and control the generation with modified text or image prompts. To enable arbitrary image editing, InstructPix2Pix~\cite{brooks2023instructpix2pix} proposed to finetune the LDM to follow user instructions that precisely convey the user intents, \textit{e.g.}, ``change the cat to dog''. The model is trained by a synthetic dataset consisting of triplets $\langle image_{src}, instruction, image_{tgt} \rangle$. The resulting model could allow both realistic and generated images and support arbitrary language instructions. Some other works also exploit a similar way to train the model to follow instructions~\cite{yu2023scaling,gu2023biomedjourney}. To further enhance the language understanding, MGIE~\cite{fu2023guiding} exploited a large multimodal model to produce a more comprehensive textual context for the instructed image editing.

In this work, we employ the image-to-image translation pipeline and are the first to formulate a GIF generation as an image translation problem. Given a reference image, the goal is to generate a realistic \emph{future} frame following a textual instruction. Therefore, the focus is on how to perform temporal rather than spatial editing on a source image. When the process rolls out, it gradually gives a sequence of frames.

\paragraph{Conditioned Video Generation.} Speaking of the high-level goal, our work resembles conditioned video generation. For video generation, a conventional way is inflating the 2D U-Net used in LDM to 3D U-Net~\cite{3dunet} by replacing the 2D convolution layers with 3D ones. Likewise, a similar strategy is taken in~\cite{ge2023preserve,he2023latent,blattmann2023align,singer2022makeavideo,wang2023lavie}, but with a slight difference in that they use interleaved spatial and temporal attention layers in the U-Net.  Due to the high cost of generating a sequence of video frames in one shot, the output videos usually have a resolution as low as $64 \times 64$. To attain high-resolution videos, these methods need to use one or more super-resolution diffusion models~\cite{li2022srdiff,saharia2022photorealistic} to upscale the resolution by 4 or 8 times. To accelerate the training, a pre-trained text-to-image LDM is usually used to initialize these models. Adding spatial-temporal modules is also a commonly used strategy for autoregressive models~\cite{wu2021godiva,wu2021nuwa,hong2022cogvideo,villegas2022phenaki}. Similarly, both~\cite{wu2021nuwa} and~\cite{hong2022cogvideo} exploit a pretrained autoregressive image generation model as the starting point. In~\cite{villegas2022phenaki}, however, the authors introduce and pretrained a new video encoder, which is then used to train a masked video decoder. 

Besides text-to-video generation, using images as the condition for video generation draws increasing attention. On one hand, once a text-to-video generation is trained, it can be further finetuned for image-to-video generation~\cite{blattmann2023stable}. In~\cite{esser2023structure}, the authors introduce additional structural conditions (\textit{e.g.}, depth maps) for more controllable video generation. Alternatively in~\cite{ni2023conditional}, a latent flow diffusion model is introduced for image-conditioned video generation by explicitly generating a sequence of optimal flows and masks as the guidance. On the other hand, a few concurrent works to ours directly approach image-to-video generation on top of video diffusion models~\cite{zhang2023i2vgen,xing2023dynamicrafter,wang2023dreamvideo}. All these works share a similar spirit to text-to-video generation models but add additional images as the reference.

Our method uses a diffusion model but differs from all the aforementioned methods in that we reformulate video generation as a frame-to-frame translation problem based on the history frames. As~\cite{brooks2023instructpix2pix} suggests image-to-image translation can maintain a decent visual consistency. In addition, we also introduce a motion flow magnitude as another condition to explicitly control the temporal dynamics. 

\paragraph{Future Frame Prediction.} Future frame prediction or forecasting~\cite{oliu2018folded,shi2015convolutional,villegas2018decomposing} has been a long-standing problem before the prevalence of diffusion models. It has been used as an anomaly detection approach by comparing the observed frame and the predicted ones~\cite{liu2018future,baradaran2023future} and video representation learning for various downstream tasks~\cite{fujitake2022video}. For these problems, a recurrent network such as LSTM~\cite{srivastava2016unsupervised}, ConvLSTM~\cite{oh2015actionconditional,villegas2018decomposing} or 3D-CNN~\cite{aigner2018futuregan} is usually used as the model architecture, and GAN~\cite{goodfellow2020generative} or Variational Autoencoder (VAE)~\cite{kingma2022autoencoding} is used as the learning objectives. With the emergence of VQ-VAE~\cite{oord2018neural}, the authors in~\cite{hu2022make} exploited axial transformer blocks to chain the encoder-decoder for autoregressive next-frame prediction. In~\cite{voleti2022mcvd}, the authors proposed masked conditional video diffusion to unify different tasks of video prediction, generation and interpolation. Nevertheless, all of these models are trained on domain-specific video datasets such as MovingMNIST~\cite{lee2019mutual}, CATER~\cite{girdhar2020cater} and UCF-101~\cite{soomro2012ucf101}, \textit{etc}, far from being a generic video generation model.

Our work takes inspiration from future frame prediction methods but proposes a simpler yet effective strategy by formulating it as an image-to-image translation problem. Furthermore, our model simultaneously takes image, text and motion magnitude as the guidance for better controllability. To attain a model as general as possible, we curate a new training dataset covering a wide range of domains. Without any further dataset-specific finetuning, our model achieves plausible video generation results as shown in Fig.~\ref{fig:teaser}. 
% Finally, we note that the goal of this work is not beating the other works but an exploration of autoregressive diffusion for video generation.
\section{Method}
\label{sec:method}

Our goal of generating GIFs, given an initial frame, a descriptive caption of motion, and a measure of optical flow to quantify the motion, is framed as an image-to-image translation problem based on latent diffusion. We first detail the process of constructing the dataset used for training our model in \cref{subsec:dataset}. Then we outline the underlying principles of our model and our training strategy in \cref{subsec:LDM}. Following this, we delve into the specifics of our proposed model, \ourmodel~,~explaining its various components in \cref{subsec:pix2gif}. Finally, we concentrate on the loss functions utilized to train our model in \cref{subsec:losses}.

%%%%%%%%%%%%%%%%%%%%%%%%%%%%%%%%%%%%%%%%%%%%%%%%%%%%%%%%%%%%%%%%
\subsection{Dataset}
\label{subsec:dataset}
\begin{figure}[t]
  \centering
   \includegraphics[width=0.95\linewidth]{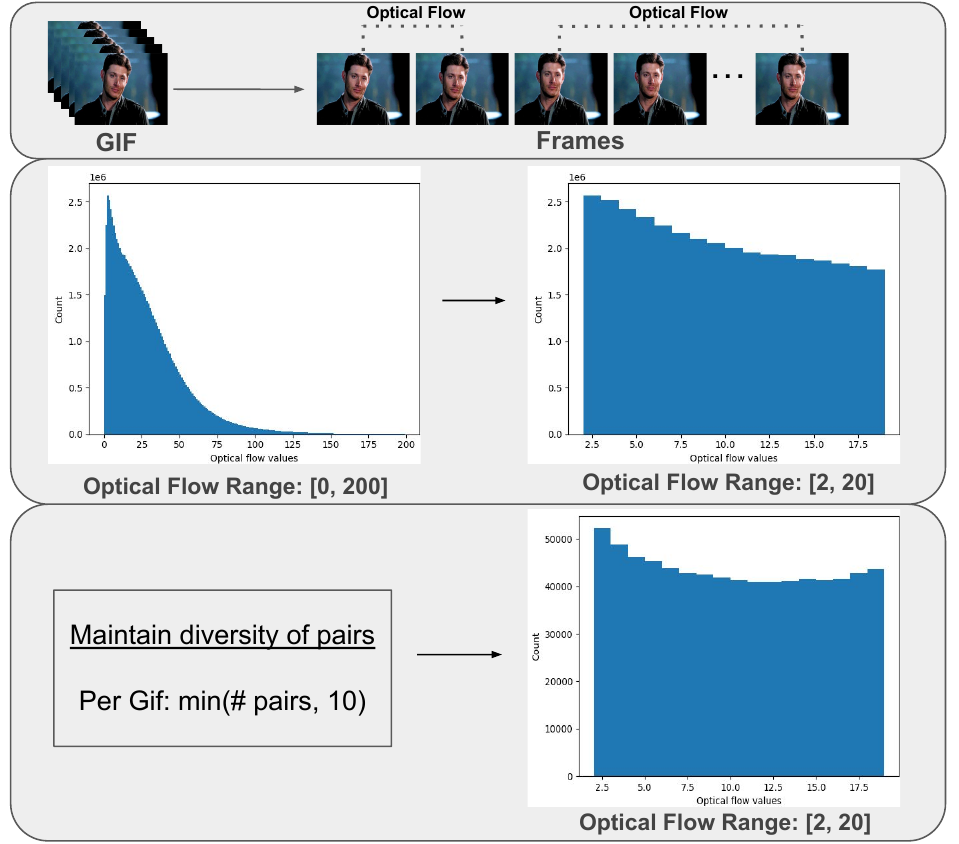}
   \caption{The three step process of curating the TGIF dataset. Starting from extracting frames to restricting the range of optical flow and then maintaining the diversity of pairs.}
   \label{fig:dataset}
\end{figure}

We utilized the Tumblr GIF (TGIF) dataset~\cite{li2016tgif}, which is primarily comprised of animated GIFs that are described by sentences or captions, displaying a preference towards content that is human-centric. The dataset comprises various types of GIFs, ranging from fast-paced to slow-paced movements, typically encapsulating a duration of 1-3 seconds. This variety ensures a broad spectrum of motion within a short time frame.

The curation process as shown in \cref{fig:dataset} involved extracting frames from all the GIFs, which exhibited varying frame rates. We then calculated the optical flow between all possible pairs of frames for a given GIF, as outlined in the Alg.~\ref{algo:flow}. The number of frames extracted from each GIF ranged from 14 to 572, with an average of approximately 41 frames. This led to a substantial number of training pairs, along with a high magnitude of optical flow for numerous pairs. The optical flow histogram calculated between all frames spans from 0 to 200. From this range, we selected the range of 2-20, which is well populated and captures smaller but significant motion. This selection rules out pairs with drastic changes caused by substantial camera motion or scene transition.

Despite restricting the range, we still obtained a significant number of training pairs, given that this range is most common. It is also possible that many GIFs might contain pairs that do not fall within this range of optical flow and for some GIFs, all its frames may be within this range. To avoid overfitting of the model on specific GIFs and to preserve diversity, we randomly selected a minimum of 10 pairs or the number of pairs within the restricted range from each GIF. This approach results in the final dataset, which ensures a nearly equal representation of all values within the selected range. In the end, the restructured dataset contains 783,184 training pairs and 105,041 validation pairs. Each data point in the dataset consists of a pair of frames from the same GIF, the corresponding GIF's caption, and the calculated optical flow between the two frames.
%%%%%%%%%%%%%%%%%%%%%%%%%%%%%%%%%%%%%%%%%%%%%%%%%%%%%%%%%%%%%%%%

\begin{algorithm}[t]
\caption{Optical flow calculation between two frames.}
\label{algo:flow}
\small
\Comment{$I_{0}$,$I_{1}$ are the frames in numpy arrays.}
  \Function{CalcOpticalFlow($I_{0}$,$I_{1}$)}{
      $I^{'}_{0}, I^{'}_{1}$ $\gets$ \FuncCall{ConvertToGrayscale}{$I_{0}$,$I_{1}$}\;
      $P_{0}$ $\gets$ \FuncCall{FeaturesToTrack}{$I^{'}_{0}$}\;
      $P_{1}$ $\gets$ \FuncCall{OpticalFlowLucasKanade}{$I^{'}_{0}$,$I^{'}_{1}$,$P_{0}$}\;
      $\var{FlowVector(F)} \gets P_{1} - P_{0}$\;
      \Comment{2D vector (x,y) to single magnitude}
      $M$ $\gets$ \FuncCall{GetMagnitude}{F}\;
      \Comment{Filter out noise and very small optical flow vectors}
      $M^{'} \gets M > \var{threshold}$\;
      \If{$M^{'} > 0$}{
        $M_{avg}$ $\gets$ \FuncCall{Average}{$M^{'}$}
        \Return{$M_{avg}$}
      }
      \Return{0}\;
  }
\end{algorithm}

% \begin{algorithm}
% \caption{Optical flow calculation between frames.}
% \label{algo:flow}
% \Comment{$I_{0}$, $I_{1}$ are the frames in numpy arrays.}
%   \Function{CalcOpticalFlow($I_{0}$, $I_{1}$)}{
%       $I^{'}_{0}, I^{'}_{1} \gets \FuncCall{ConvertToGrayscale}{I_{0}, I_{1}}$
%       % $P_{0} \gets \FuncCall{FeaturesToTrack}{I^{'}_{0}}$
%       % $P_{1} \gets \FuncCall{OpticalFlowLucasKanade}{I^{'}_{0}, I^{'}_{1}, P_{0}}$
%       % $\var{FlowVector(F)} \gets P_{1} - P_{0}$
%       % \Comment{2D vector (x,y) to single magnitude}
%       % $M \gets \FuncCall{GetMagnitude}{F}$
%       % \Comment{Filter out noise and very small optical flow vectors}
%       % $M^{'} \gets M > \var{threshold}$
%       % \If{$M^{'} > 0$}{
%       %   $M_{avg} \gets \FuncCall{Average}{M^{'}}$
%       %   \Return{$M_{avg}$}
%       % }
%       % \Return{0}\;
%   }
% \end{algorithm}
\begin{figure*}[t]
  \centering
   \includegraphics[width=0.95\linewidth]{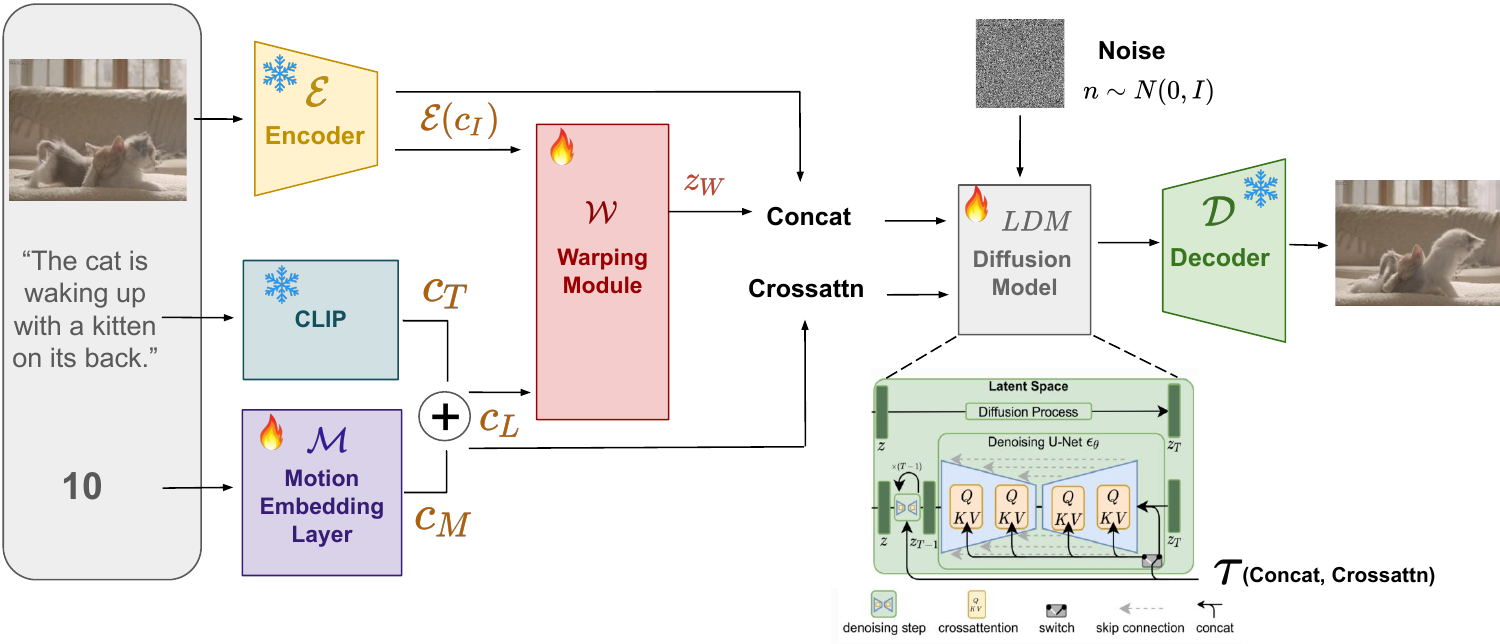}
   \caption{\textit{Pix2Gif} model pipeline. We propose an end-to-end network where the inputs are encoded by $\mathcal{E}$, CLIP and $\mathcal{M}$ to output $\mathcal{E}(c_L)$,$c_T$ and $c_M$ respectively, which then goes into $\mathcal{W}$ to form the conditioning input for $LDM$.}
   \label{fig:model}
\end{figure*}

%%%%%%%%%%%%%%%%%%%%%%%%%%%%%%%%%%%%%%%%%%%%%%%%%%%%%%%%%%%%%%%%
\subsection{Preliminary: Instructed Image Editing}
% \subsection{Text-conditioned Image Translation}
\label{subsec:LDM}
% To initiate our training, we utilize a pre-trained Stable Diffusion model, an extensive scale latent diffusion model. This training, carried out in the latent space, augments the model's efficiency and quality, thanks to the minimized dimensions and quantized features facilitated by a variational autoencoder. This autoencoder is composed of a pre-trained encoder $\mathcal{E}$ which transforms the images from pixel space to latent space.

Our model is fundamentally grounded in the latent diffusion models (LDMs) for image generation and editing~\cite{rombach2022highresolution,brooks2023instructpix2pix}. More specifically, we build upon InstructPix2Pix~\cite{brooks2023instructpix2pix} by framing our objective in the context of an instructed image-to-image translation task. Given an image $x$, the forward diffusion procedure introduces noise to the encoded latent $z$, thereby producing a noisy latent vector $z_t$. This process is carried out over $T$ timesteps, with each timestep $t \in \{1,...,T\}$ seeing an increment in the noise level until it culminates into a random noise $n$. A network $e_{\theta}$ is trained by minimizing the following latent diffusion objective to predict noise existing in the noisy latent $z_{t}$, considering factors image conditioning $c_{I}$ and textual instruction $c_{T}$:
\begin{equation}
  L_{LDM} = \mathbb{E}_{\psi}\left[||\epsilon-\epsilon_{\theta}(z_{t},t,\mathcal{E}(c_{I}),c_{T})||^2_2\right]
  \label{eq:ldm_obj_orig}
\end{equation}
\begin{equation}
  \Psi = \mathcal{E}(x),\mathcal{E}(c_{I}),c_{T}, \epsilon\sim\mathcal{N}(0,1),t
  % \label{eq:ldm_obj}
\end{equation}
where $\mathcal{E}$ is the VQ-VAE encoder that transforms the images from pixel space to discrete latent space. To facilitate image conditioning, $z_t$ and $\mathcal{E}(c_{I})$ are concatenated and then fed into a convolutional layer. The model is trained for conditional and unconditional denoising, given the image and caption condition individually or collectively.

%%%%%%%%%%%%%%%%%%%%%
% We build upon the foundation of the InstructPix2Pix ~\cite{brooks2023instructpix2pix} codebase to frame our objective in the context of an image-to-image editing task.
% For conditioning, we employ a mechanism similar to~\cite{brooks2023instructpix2pix}, where we utilize a linear addition of caption conditioning and motion conditioning as the input for text conditioning. To facilitate learning from the warped latent $z_W$, in addition to the image conditioning, we concatenate $z_t$, $z_W$, and $c_I$. We continue to utilize the identical score function as implemented by ~\cite{brooks2023instructpix2pix}, with no modifications. This sustains our model's capacity to learn conditional or unconditional denoising, given the image and caption condition individually or collectively. 
%%%%%%%%%%%%%%%%%%%%%%%%%%%%%%%%%%%%%%%%%%%%%%%%%%%%%%%%%%%%%%%

\subsection{Our Model: Pix2Gif}
\label{subsec:pix2gif}

We build our model similar to InstructPix2Pix and frame our objective in the context of a text-instructed and motion-guided temporal editing problem. 
% For conditioning, we employ a mechanism similar to~\cite{brooks2023instructpix2pix}, where we utilize a linear addition of caption conditioning and motion conditioning as the input for text conditioning. 
% To facilitate learning from the warped latent $z_W$, in addition to the image conditioning, we concatenate $z_t$, $z_W$, and $c_I$. 
% We continue to utilize the identical score function as implemented by~\cite{brooks2023instructpix2pix}, with no modifications. This sustains our model's capacity to learn conditional or unconditional denoising.
% , given the image and caption condition individually or collectively. 
% The fundamental concept of our model is to generate subsequent frames from a given image and text by manipulating the input of the optical flow. 
Compared with the original InstructPix2Pix pipeline, the main innovation is the newly introduced motion-based warping module. The overall model pipeline is shown in Fig.~\ref{fig:model}.

Our model takes three inputs: an image, a text instruction, and a motion magnitude. These inputs are fed into the model through two pathways - once through the diffusion model directly, and again through the warping module, which will be discussed in \cref{subsubsec:motion_embed} and \cref{subsubsec:warping}. When passed through the caption, we add the phrase ``The optical flow is \textunderscore.'' to the original caption. The flow input is then appended at the end in a word format rather than a numerical one, as the CLIP model tends to assign higher similarity scores to word forms than to numerical representations of numbers for the same image. Finally, our model is trained by minimizing the following loss function: 
% Given an image $x$, the forward diffusion procedure introduces noise to the encoded latent $z$, thereby producing a noisy latent vector $z_t$. This process is carried out over $T$ timesteps, with each timestep $t \in T$ seeing an increment in the noise level until it culminates into a random noise $n$. A network $e_{\theta}$ is trained by minimizing the following latent diffusion objective to predict noise existing in the noisy latent $z_{t}$, considering factors like image conditioning $c_{I}$, caption conditioning $c_{T}$, and motion conditioning $c_{M}$.
\begin{equation}
  L'_{LDM} = \mathbb{E}_{\psi}\left[||\epsilon-\epsilon_{\theta}(z_{t},t,\mathcal{E}(c_{I}),c_{T},c_{M})||^2_2\right]
  \label{eq:ldm_obj}
\end{equation}
\begin{equation}
  \Psi = \mathcal{E}(x),\mathcal{E}(c_{I}),c_{T},c_{M},\epsilon\sim\mathcal{N}(0,1),t
  % \label{eq:ldm_obj}
\end{equation}
where $c_{M}$ is the motion condition. The altered caption is processed via the pre-trained CLIP model to yield $c_{T}$, while the output of $\mathcal{M}$ gives us $c_{M}$. These two conditions are then added linearly, serving as the conditioning input for both the Warping Module $\mathcal{W}$ (discussed in \cref{subsubsec:warping}) and the Latent Diffusion Model $LDM$ (referenced in \cref{subsec:LDM}).

\subsubsection{Motion Embedding Layer}
\label{subsubsec:motion_embed}
% \begin{itemize}
%     \item brief para
%     \item Give the intuition of using this layer
%     \item for example, inputting the motion information externally
%     \item and using it in the warping module
% \end{itemize}
In conventional conditional diffusion models, text conditioning or prompts are usually sufficient to generate the desired images or edits, as in the case of ~\cite{brooks2023instructpix2pix}. Initially, we indirectly passed the motion input through the prompt. However, this approach resulted in the model giving divided attention to a single token in the caption, which would have been acceptable under normal circumstances. But in our case, the main caption often remains the same while the motion input changes to generate subsequent frames. 
% \begin{equation}
%     c_L = c_T + c_M
% \end{equation}
To enable the model to focus primarily and independently on the motion input, we incorporate a simple embedding layer $\mathcal{M}$. This layer converts the motion input into an integer and selects an embedding vector from the learned embedding matrix. This vector is then duplicated and concatenated with itself to generate $c_M$, which when combined with the caption embedding $c_T$, provides the conditioning input $c_L = c_T + c_M$ for both warping module $\mathcal{W}$ and $LDM$.

\subsubsection{Warping Module}
\label{subsubsec:warping}
% \begin{itemize}
%     \item Fig for warping module
%     \item write about the inputs
%     \item Describe flownet
%     \item how the cond is incorporated in the flownet
% \end{itemize}

\begin{figure}[t]
  \centering
   \includegraphics[width=0.95\linewidth]{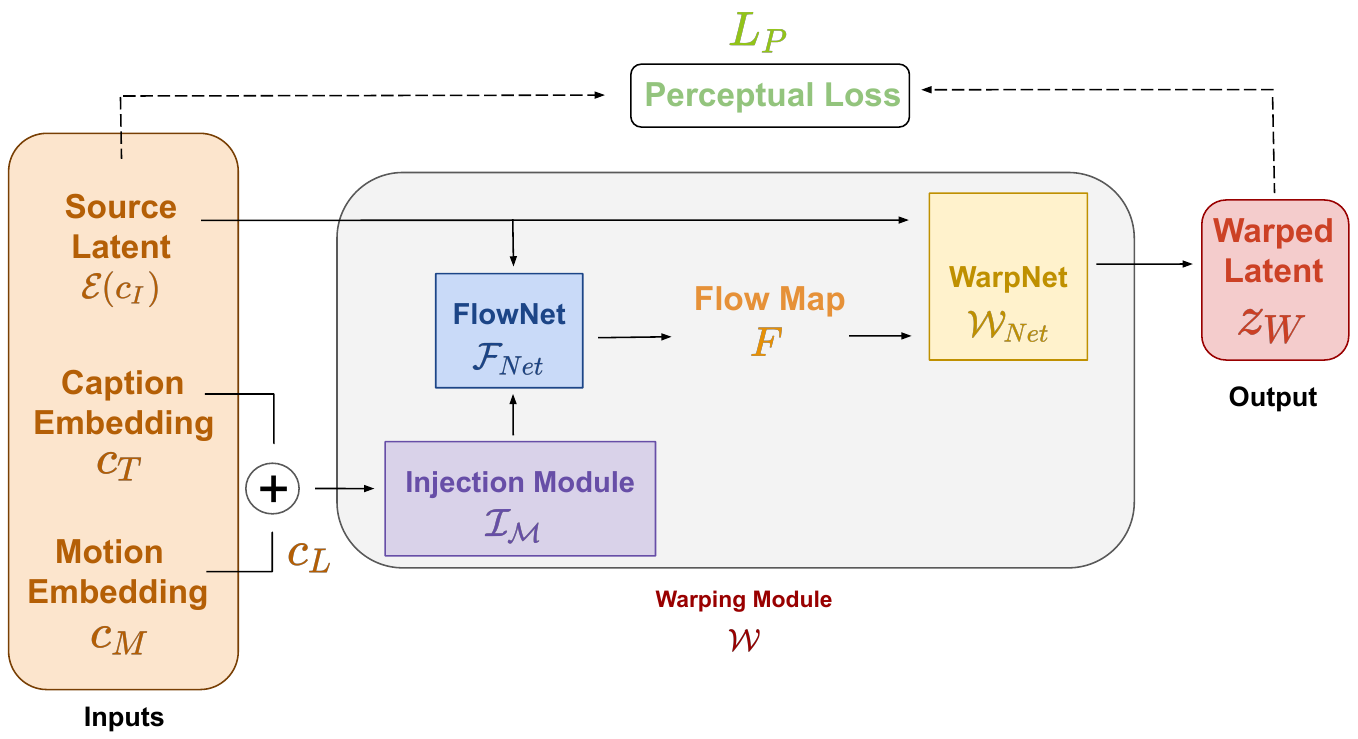}

   \caption{Deep dive into the Warping Module $\mathcal{W}$. It comprises of three units: $\mathcal{I}_M$, $\mathcal{F}_{Net}$ and $\mathcal{W}_{Net}$.}
   \label{fig:warp}
\end{figure}

One of the main components of Pix2Gif is the Warping Module $\mathcal{W}$. As illustrated in \cref{fig:warp}, it technically comprises two networks: the FlowNet ($\mathcal{F}_{Net}$) and the WarpNet ($\mathcal{W}_{Net}$). Ordinarily, the computation of optical flow involves two images. However, in this case, we initially have only one image - the source image - and that too in the latent domain. Thus, our goal is to learn the optical flow utilizing just one latent image. This is achieved via $\mathcal{F}_{Net}$, conditioned on $c_L$, which guides it to generate a flow feature map in the intended direction with the hint of text and motion prompts. This condition is processed by the Injection Module ($\mathcal{I_{M}}$), a compact encoder designed to make $c_L$ compatible for concatenation with one of the intermediate feature maps near the end of the network. This configuration enables $\mathcal{F}_{Net}$ to independently learn high-level features, which are then guided in the desired direction with the introduction of $c_L$.
\begin{equation}
    F = \mathcal{F}_{Net}(\mathcal{E}(c_I), \mathcal{I_M}(c_L))
\end{equation}
\begin{equation}
    z_W = \mathcal{W}_{Net}(\mathcal{E}(c_I), F)
\end{equation}
The architecture of $\mathcal{F}_{Net}$ resembles that of UNet, producing an output with a fixed channel of two to capture changes in the horizontal and vertical components. 

This optical flow feature map ($F$), along with the source latent ($\mathcal{E}(c_I)$), is then processed through $\mathcal{W}_{Net}$ to yield a Fischer map $z_W$ of $\mathcal{E}(c_I)$. This transformation is learned more efficiently and abstractly in the latent space than in the pixel space.

%%%%%%%%%%%%%%%%%%%%%%%%%%%%%%%%%%%%%%%%%%%%%%%%%%%%%%%%%%%%%%
\subsection{Losses}
\label{subsec:losses}
% \begin{itemize}
%     \item One we have the normal L2 loss between the source image and the target image.
%     \item This is mostly a reconstruction loss
%     \item Second we have the perceptual loss between the source image latents and the warped source image latents.
%     \item This is to ensure that the warped image maintains the high level features (e.g., edges, textures, object types), resulting in images that look more perceptually and semantically similar to human eyes and preserves the overall structure of the source image in addition to just pixels.
%     \item Third loss is the smoothness loss applied to the flow feature map.
%     \item This is a regularizer which ensures that there is no abrupt and unrealistic changes in motion of neighboring pixels.
% \end{itemize}
Our model incorporates two different types of losses. The first type is the standard L2 loss \cref{eq:ldm_obj}, which is utilized by the stable diffusion model and talked about in \cref{subsec:LDM}
   
The second type of loss incorporated in our model is the perceptual loss. This is calculated by comparing the latent features of the image condition $\mathcal{E}(c_{I})$ and those of the warped image $z_W$. To implement the perceptual loss, a pre-trained VGG network~\cite{simonyan2015deep} is used, but with a modification to its input layer to accommodate 4 channels instead of the standard 3. This modification is realized by averaging the weights from the first three channels and using this average to initialize the fourth channel. Given both the latent feature maps, the perceptual loss $L_p$ can be calculated using the pre-trained modified VGG network as follows. Let $\phi_{k}$($\mathcal{E}(c_{I})$) and $\phi_{k}$($z_W$) be the feature maps of the k-th layer of the VGG network when $\mathcal{E}(c_{I})$ and $z_W$ are forward propagated through it. The perceptual loss $L_p$ is then defined as:
\begin{equation}
    L_p(\mathcal{E}(c_{I}), z_W) = \sum_{k} \lambda_{k} || \phi_{k}(\mathcal{E}(c_{I})) - \phi_{k}(z_W) ||^2
\end{equation}
Here, $||.||$ denotes the Frobenius norm, and $\lambda_k$ is a weighting factor that balances the contribution of each layer to the total perceptual loss. Each layer k in the VGG network captures different levels of abstraction in the image, and the perceptual loss ensures that these abstractions are similar for both images.
The purpose of this loss is to ensure that the warped image retains the high-level features such as edges, textures, and object types, resulting in images that are more perceptually and semantically similar to the human eye. In addition to pixel-level fidelity, it also helps preserve the overall structure of the source image. Perceptual loss considers the perceptual and semantic differences between the reconstructed and original image rather than just pixel-level differences.  
   
% The third loss is the smoothness loss, which is applied to the flow feature map. This loss serves as a regularizer, ensuring there are no abrupt and unrealistic changes in the motion of neighboring pixels. Essentially, the smoothness loss encourages spatial consistency in the flow field, leading to smoother and more coherent motion transitions. Given a flow map $F$ of dimensions $(N, C, H, W)$ where $N$ is the batch size, $C$ is the number of channels which for $F$ is 2, $H$ is the height, and $W$ is the width, the smoothness loss $L_S$ can be calculated as:
% \begin{equation}
% \begin{aligned}
%     \label{eq:L_S}
%     L_S(F) &= \frac{1}{NCHW} \sum_{n=1}^{N} \sum_{c=1}^{C} \sum_{h=1}^{H} \sum_{w=2}^{W} |F_{nchw}-F_{nchw-1}|\\
%     &\quad + \frac{1}{NCHW} \sum_{n=1}^{N} \sum_{c=1}^{C} \sum_{h=2}^{H} \sum_{w=1}^{W} |F_{nchw}-F_{nch-1w}|
% \end{aligned}
% \end{equation}

% Here, $|\cdot|$ denotes the absolute value. The first term represents the mean absolute difference between neighboring elements in the width direction, and the second term represents the mean absolute difference between neighboring elements in the height direction. The smoothness loss encourages the flow to be smooth by penalizing large differences between neighboring flow vectors.

In conclusion, the total loss function, denoted as $L_{T}$, for our objective is computed by a weighted sum of the two individual losses.

\begin{equation}
    L_{T} = L'_{LDM} + \lambda_{P}L_{P} %+ \lambda_{S}L_{S}
\end{equation}

Here, $\lambda_{P}$ is the weighting factor for perceptual loss. These two losses together provide a holistic framework to train our model by ensuring pixel-level accuracy, preservation of high-level features, and smooth motion transitions.
\section{Experiments}
\label{sec:experiments}

% \begin{figure*}
%   \centering
%    \includegraphics[width=0.6\linewidth]{images/CVPR_dog.pdf}
%    \end{figure*}
%    \begin{figure*}
%    \centering
%   \includegraphics[width=0.6\linewidth]{images/CVPR_horse.pdf}
%   % \caption{\jw{we need to change images that can show clear differences across different methods.}}
%     \end {figure*}
%     \begin{figure*}    
%     \centering
%   \includegraphics[width=0.6\linewidth]{images/CVPR_joker.pdf}
%    \caption{Comparison studies with InstructPix2Pix~\cite{brooks2023instructpix2pix}, our Base Model and our Proposed model Pix2Gif.
%    Given the source image (in the 1st column), subsequent frames are generated at 256 resolution. \jw{add an illustration in the figure as well to tell readers that the first column is the input frame and the others are generated frames. Make this figure bigger.}\textcolor{blue}{done}}
%    \label{fig:viz}
% \end{figure*}

\begin{figure*}
\centering
\begin{subfigure}[b]{0.95\textwidth}
   \includegraphics[width=1\linewidth]{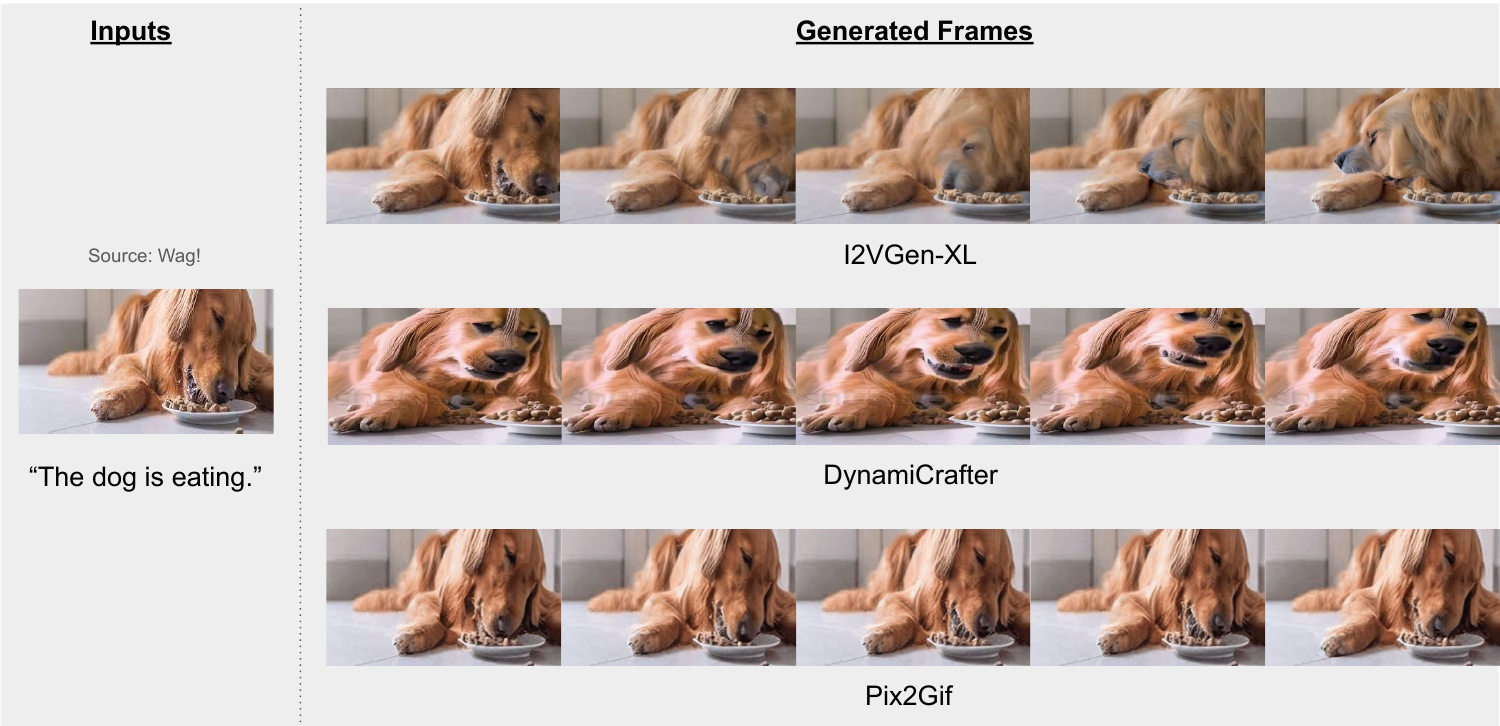}
   \caption{}
   \label{fig:dog} 
\end{subfigure}

\begin{subfigure}[b]{0.95\textwidth}
   \includegraphics[width=1\linewidth]{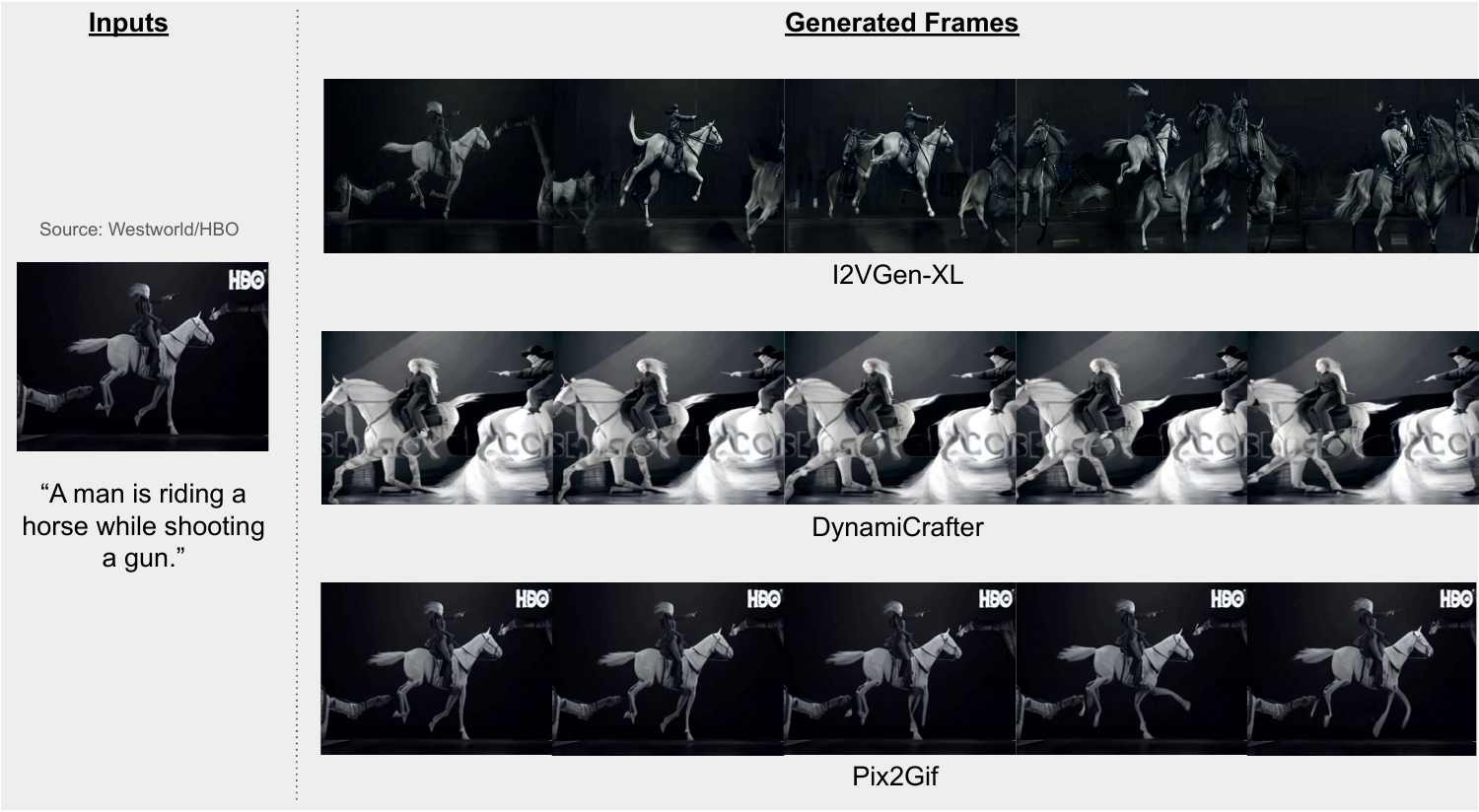}
   \caption{}
   \label{fig:horse}
\end{subfigure}
\caption{Comparison studies with other image-text to video models.
Given a source image and a caption, frames are extracted from the generated 16-frame video at 256x256 resolution.}
\label{fig:viz}
\end{figure*}

\begin{table*}
\begin{center}
\scalebox{0.95}{
\begin{tabular}{@{}lccccccl@{}}\toprule
\multicolumn{1}{c}{\multirow{2}{*}{\textbf{Method}}} & \multicolumn{3}{c}{\textbf{UCF-101~\cite{soomro2012ucf101}}} & \multicolumn{3}{c}{\textbf{MSR-VTT~\cite{xu2016msr}}} \\ 
\cmidrule(lr){2-4}\cmidrule(lr){5-7}
\multicolumn{1}{c}{}                                  & \textbf{FVD} $\downarrow$  & \textbf{CLIPSim} $\uparrow$ & \textbf{PIC} $\uparrow$  & \textbf{FVD} $\downarrow$  & \textbf{CLIPSim} $\uparrow$ & \textbf{PIC} $\uparrow$  \\\midrule 
I2VGen-XL~\cite{zhang2023i2vgen}      & 563.12       & \textbf{0.2865}    & 0.6329 & 278.62       & 0.2272    & 0.6018 \\
DynamiCrafter~\cite{xing2023dynamicrafter}   & 527.06       & 0.2796    & 0.6307 & 271.63       & \textbf{0.2602}    & 0.6135     \\% \midrule
Pix2Gif~(Ours)         & \textbf{285.02}        & 0.2815    & \textbf{0.8763}   & \textbf{168.69}        & 0.2573    & \textbf{0.8521} \\\bottomrule
\end{tabular}}
\caption{Quantitative comparison with state-of-the-art image-text-to-video generation models for the zero-shot setting.}
\label{tab:fvd}\end{center}
\vspace*{-0.5mm}
\end{table*}

\subsection{Setup}

\paragraph{Datasets} We utilize the Tumblr GIF (TGIF) dataset for our training and validation purposes as discussed in \cref{subsec:dataset}. We evaluate our model on two datasets: MSR-VTT~\cite{xu2016msr} and UCF-101~\cite{soomro2012ucf101}, following the common practice. For these datasets, we follow the sampling strategy as outlined in \cite{xing2023dynamicrafter}.

\paragraph{Implementations} Our model is initialized with the exponential moving average (EMA) weights of the Stable Diffusion v1.5 checkpoint\footnote{https://huggingface.co/runwayml/stable-diffusion-v1-5/blob/main/v1-5-pruned-emaonly.ckpt} and the improved ft-MSE autoencoder weights\footnote{https://huggingface.co/stabilityai/sd-vae-ft-mse-original/blob/main/vae-ft-mse-840000-ema-pruned.ckpt}. We trained the model at 256x256 resolution for 7 epochs on a single node of 16 V100 GPUs for 25k steps. We used the AdamW optimizer with a learning rate of $10^{-4}$. We set the weighting factor for perceptual loss ($\lambda_{P}$) as $10^{-2}$.

\paragraph{Metrics} We report Frechet Video Distance (FVD) \cite{unterthiner2019fvd}, CLIP Similarity (CLIPSim) which is the average similarity calculated for all the generated frames with the input caption and Perceptual Input Conformity (PIC) as described in \cite{xing2023dynamicrafter} for all methods. For comparison, we assess the zero-shot generation performance on I2VGen-XL~\cite{zhang2023i2vgen} and DynamiCrafter~\cite{xing2023dynamicrafter}. 

\begin{figure*}
\centering
\begin{subfigure}{.32\textwidth}
  \centering
  \animategraphics[width=\textwidth, loop, autoplay]{6}%frame rate
    {images/figures/all-cat_per_new_2_e3_cat-is-playing-with-wool_100_1371_7_gif_frame-}%path to figures
    {0}%start index
    {7}%end index
  \caption{Action1: Cat is playing with wool}
  \label{fig:action1}
\end{subfigure}\hspace{2mm}%
\begin{subfigure}{.32\textwidth}
  \centering
  \animategraphics[width=\textwidth, loop, autoplay]{6}%frame rate
    {images/figures/all-cat_per_new_2_e3_cat-is-dancing-to-the-music_100_1371_7_gif_frame-}%path to figures
    {0}%start index
    {7}%end index
  \caption{Action2: Cat is dancing}
  \label{fig:action2}
\end{subfigure}\hspace{2mm}%
\begin{subfigure}{.32\textwidth}
  \centering
  \animategraphics[width=\textwidth, loop, autoplay]{6}%frame rate
    {images/figures/all-cat_per_new_2_e3_cat-is-dancing-to-the-music-while-playing-with-wool_100_1371_7_gif_frame-}%path to figures
    {0}%start index
    {7}%end index
  \caption{Action1+2: Cat dancing and playing with wool}
  \label{fig:composition}
\end{subfigure}
\begin{subfigure}{.3\textwidth}
  \centering
  \includegraphics[width=\linewidth]{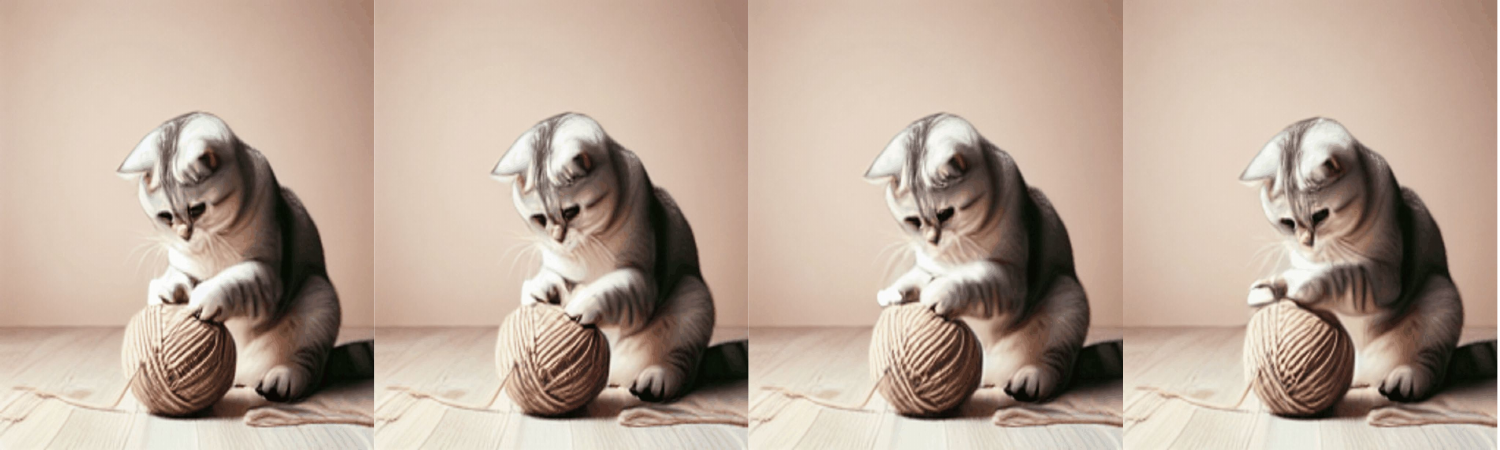}
  % \caption{with the source frame}
  % \label{fig:abl_src}
\end{subfigure}\hspace{4mm}%
\begin{subfigure}{.3\textwidth}
  \centering
  \includegraphics[width=\linewidth]{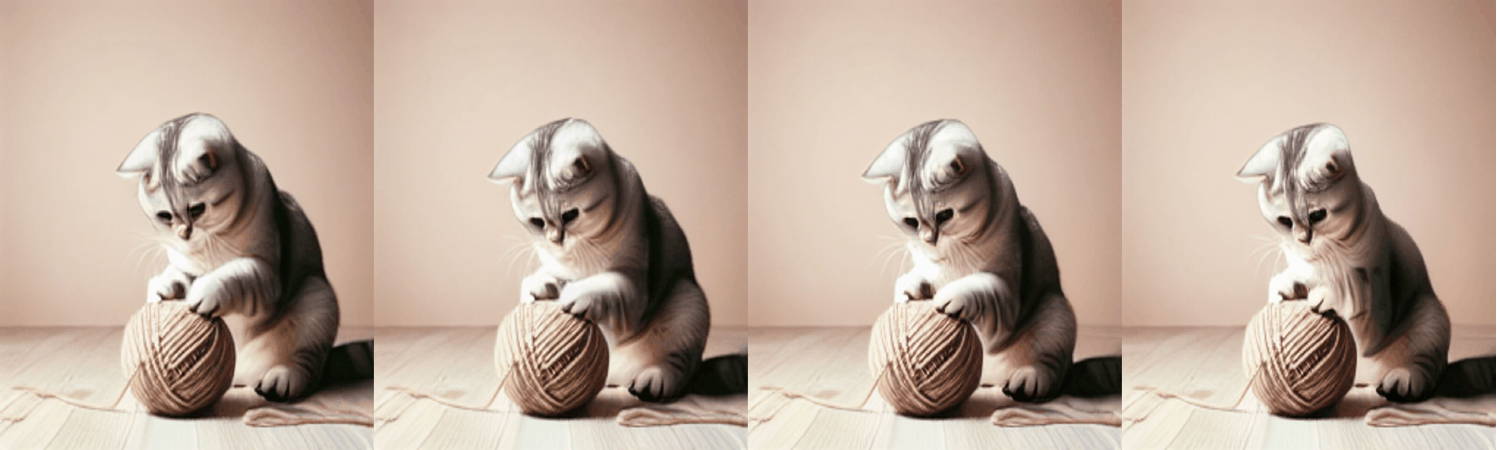}
  % \caption{with the target frame}
  % \label{fig:abl_tgt}
\end{subfigure}\hspace{4mm}%
\begin{subfigure}{.3\textwidth}
  \centering
  \includegraphics[width=\linewidth]{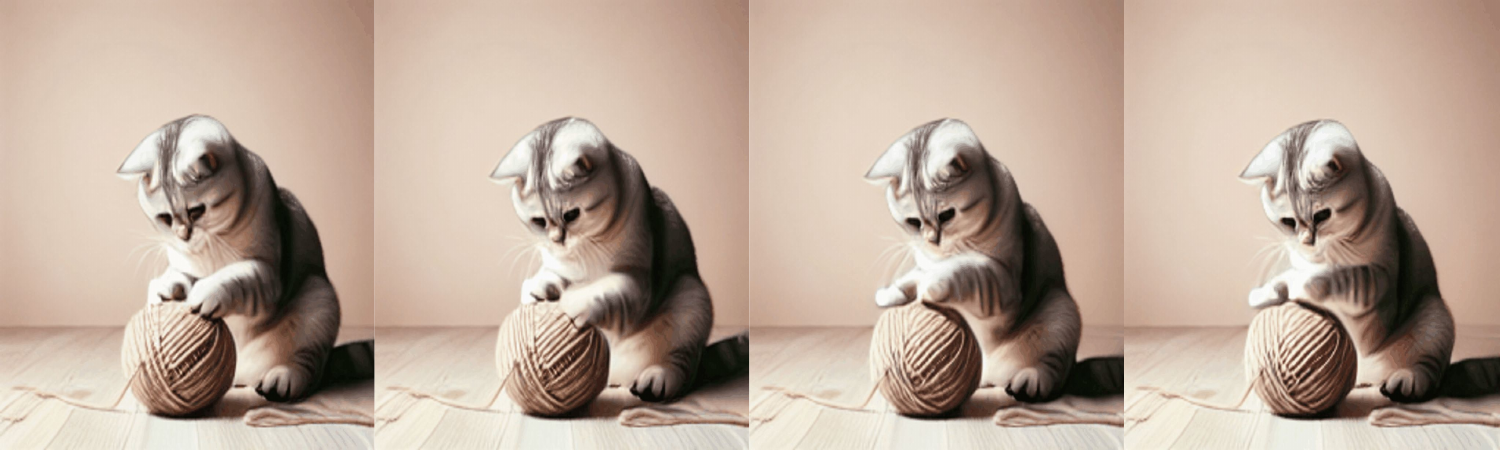}
  % \caption{with the source frame}
  % \label{fig:abl_src}
\end{subfigure}%

\caption{\textit{Pix2Gif} showing composition capabilities for different types of motions. [GIFs best viewed in Adobe Acrobat Reader]}
\label{fig:compositionality}
\end{figure*}

\begin{table*}
\begin{center}
\scalebox{0.95}{
\begin{tabular}{@{}lccccccccl@{}}\toprule
\multicolumn{1}{c}{\multirow{2}{*}{\textbf{Method / cfg$\_$img}}} & \multicolumn{2}{c}{\textbf{1.4}} & \multicolumn{2}{c}{\textbf{1.6}} & \multicolumn{2}{c}{\textbf{1.8}} & \multicolumn{2}{c}{\textbf{2.0}} \\ 
\cmidrule(lr){2-3}\cmidrule(lr){4-5}\cmidrule(lr){6-7}\cmidrule(lr){8-9}
\multicolumn{1}{c}{}                                  & \textbf{L2} $\downarrow$  & \textbf{PCC} $\uparrow$ & \textbf{L2} $\downarrow$  & \textbf{PCC} $\uparrow$  & \textbf{L2} $\downarrow$ & \textbf{PCC} $\uparrow$ &
\textbf{L2} $\downarrow$ & \textbf{PCC} $\uparrow$
\\\midrule 
InstructPix2Pix~\cite{brooks2023instructpix2pix}       & 23.429       & -0.229    & 25.492 & -0.028       & 27.037    & -0.423 & 27.530 & 0.139 \\
Pix2Gif-Base   & 7.580       & 0.989    & 5.188 & 0.987       & 5.595    & 0.992 & 7.029 & 0.991     \\% \midrule
Pix2Gif         & \textbf{1.746}        & \textbf{0.995}    & \textbf{1.972}   & \textbf{0.995}        & \textbf{2.944}    & \textbf{0.997} & \textbf{4.076} & \textbf{0.997} \\\bottomrule
\end{tabular}}
\caption{Ablation study comparing image translation methods with a focus on motion coherency at varying cfg$\_$img values.}
\label{tab:coherency}
\end{center}
\end{table*}

\subsection{Results}

\paragraph{Comparisons with previous works} 

\cref{fig:viz} and \cref{tab:fvd} provide a qualitative and quantitative comparison of three image-text to video models: I2VGen-XL~\cite{zhang2023i2vgen}, DynamiCrafter~\cite{xing2023dynamicrafter}, and our \textit{Pix2Gif}. 

In \cref{fig:dog}, the I2VGen-XL model misshapes the dog's face and generates it sideways in a nonsensical manner. DynamiCrafter appears to disregard the input parameters, as the initial frame differs significantly in position, color, and texture. It is also challenging to discern whether the dog is eating or merely moving its mouth. Our model, \textit{Pix2Gif}, accurately retains all the dog's details and successfully depicts it eating from a plate. In \cref{fig:horse}, we assess the models' capabilities by generating a video from a relatively dark image. Once again, I2VGen-XL starts strong, producing some impressive frames, but these soon turn into highly stylized and improbable images. DynamiCrafter appears to misinterpret the input image, generating something significantly different, although it seems to adhere to the caption. Conversely, \textit{Pix2Gif} comprehends the inputs effectively and produces corresponding motion while preserving the overall integrity of the source image.
   
% In \cref{fig:joker}, I2VGen-XL depicts the Joker laughing, but there are numerous distortions in elements like the Joker's purple coat and teeth in later frames. DynamiCrafter does a decent job with the face but struggles to maintain the coat and hand's structure. In contrast, Pix2Gif successfully retains the face and all accessories, making the Joker smile and move his head simultaneously.  
   
Quantitatively, \textit{Pix2Gif} excels in both the FVD and PIC metrics shown in Tab.~\ref{tab:fvd}, which aligns with our observations of the frames generated in \cref{fig:viz}. These frames effectively preserve the structure and closely adhere to the input prompts (source image and caption). However, \textit{Pix2Gif} does not perform as well in the CLIPSim metric, despite accurately following the caption. The other two models as seen in \cref{fig:viz} do follow the caption, but they fail to adhere to the input image and produce plausible temporal transitions. This is partially attributed to the inherent model design in these two methods. Both methods attempt to generate a full sequence of frames at once using the 3D diffusion network, which inevitably compounds the spatial and temporal dimensions. Moreover, the results indicate that they function more as text-to-video models than image-text-to-video models, especially DynamiCrafter. This discrepancy also raises questions about the effectiveness of the CLIPSim metric for evaluating image-text-to-video models and calls for more sophisticated metrics for evaluating video generation.

\paragraph{Compositionality of actions} \cref{fig:compositionality} illustrates an intriguing emerging capability of \textit{Pix2Gif}: the ability to combine actions. In \cref{fig:action1}, we see a cat playing with wool, with only the cat's paws and the wool moving. In \cref{fig:action2}, we instruct the cat to dance, resulting in the cat moving its body but the wool remaining still. Finally, in \cref{fig:composition}, we provide a caption that blends the actions from \cref{fig:action1} and \cref{fig:action2}. The result is a scene where the cat is both moving the wool and its body. This demonstrates \textit{Pix2Gif}'s ability to comprehend the caption and its associated motion, and to convert that understanding into a GIF. Such compositional capability significantly increases user controllability, a crucial aspect for practical applications.

\begin{figure*}
\centering
\begin{subfigure}{.24\textwidth}
  \centering
  \includegraphics[width=\linewidth]{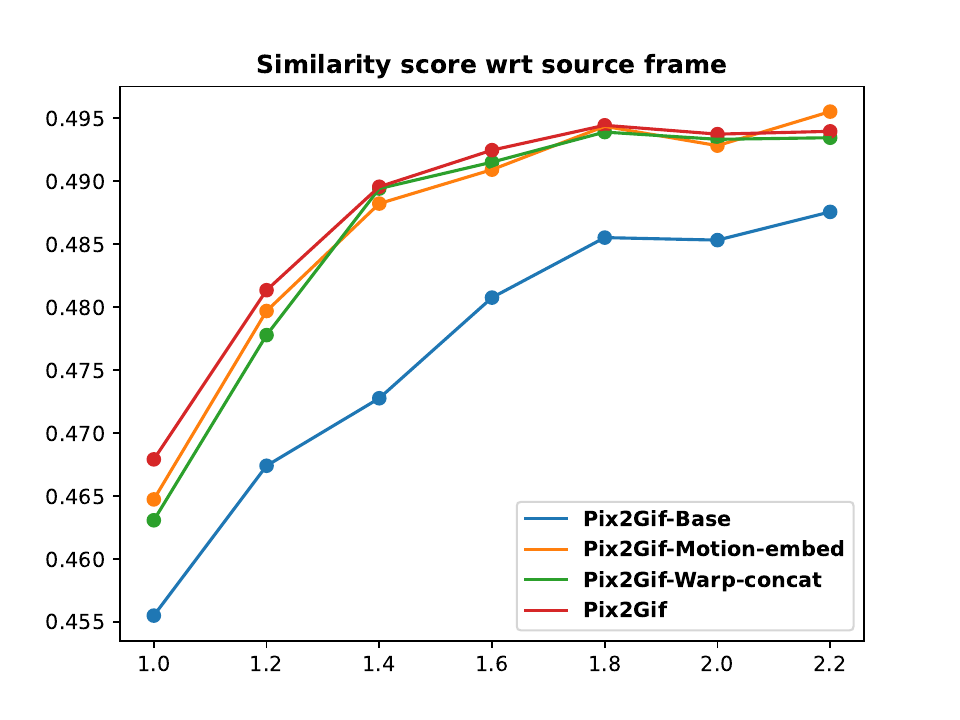}
  \caption{Source frame score}
  \label{fig:abl_src}
\end{subfigure}%
\begin{subfigure}{.24\textwidth}
  \centering
  \includegraphics[width=\linewidth]{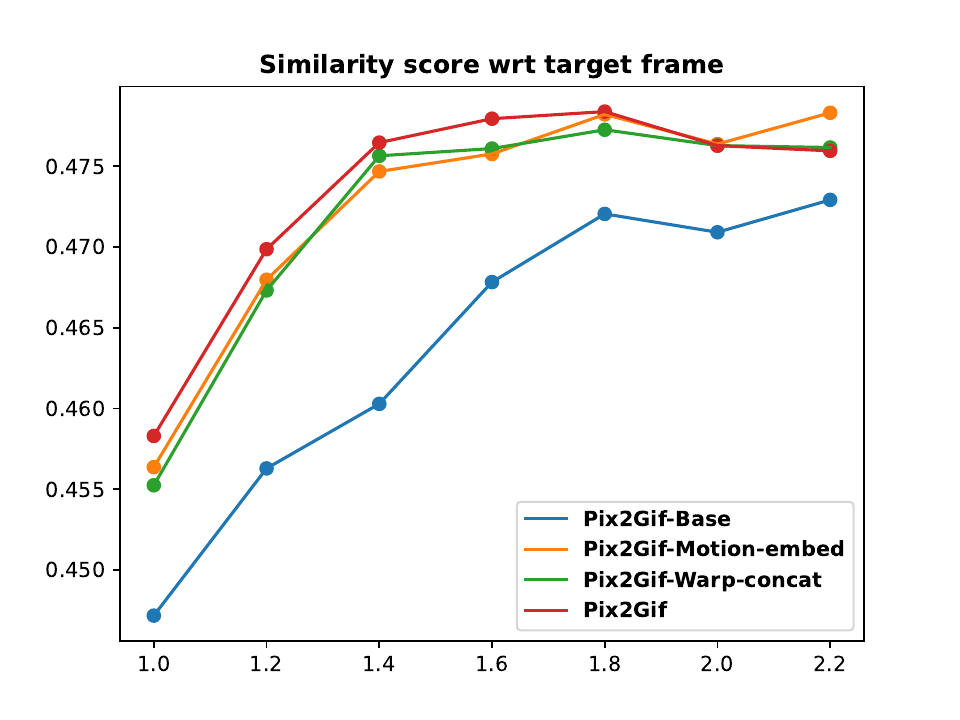}
  \caption{Target frame score}
  \label{fig:abl_tgt}
\end{subfigure}
\begin{subfigure}{.24\textwidth}
  \centering
  \includegraphics[width=\linewidth]{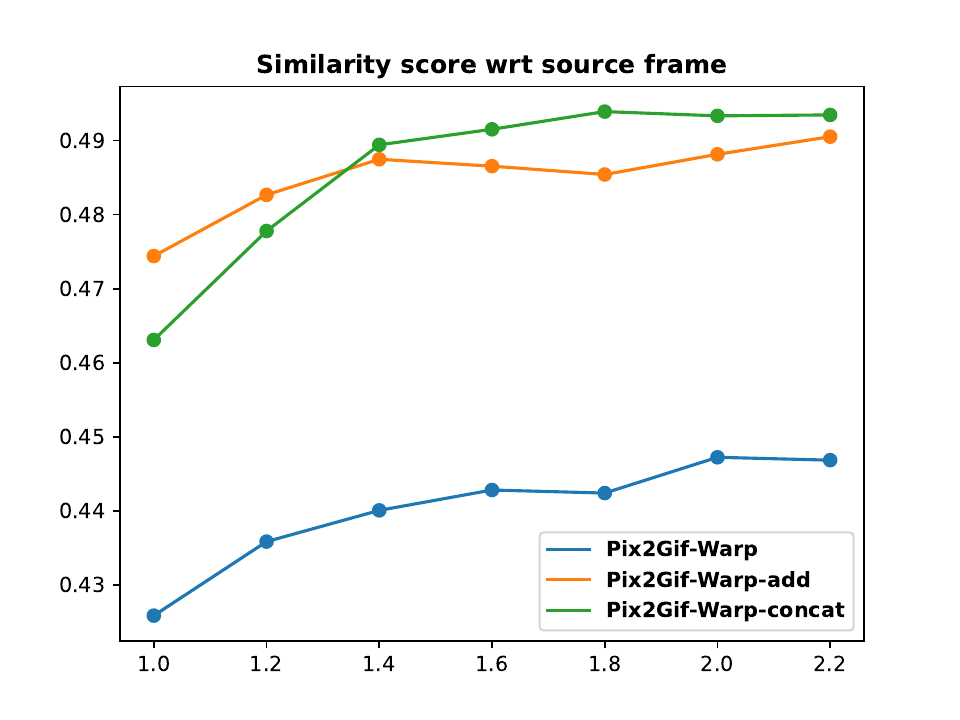}
  \caption{Source frame score}
  \label{fig:warp_src}
\end{subfigure}%
\begin{subfigure}{.24\textwidth}
  \centering
  \includegraphics[width=\linewidth]{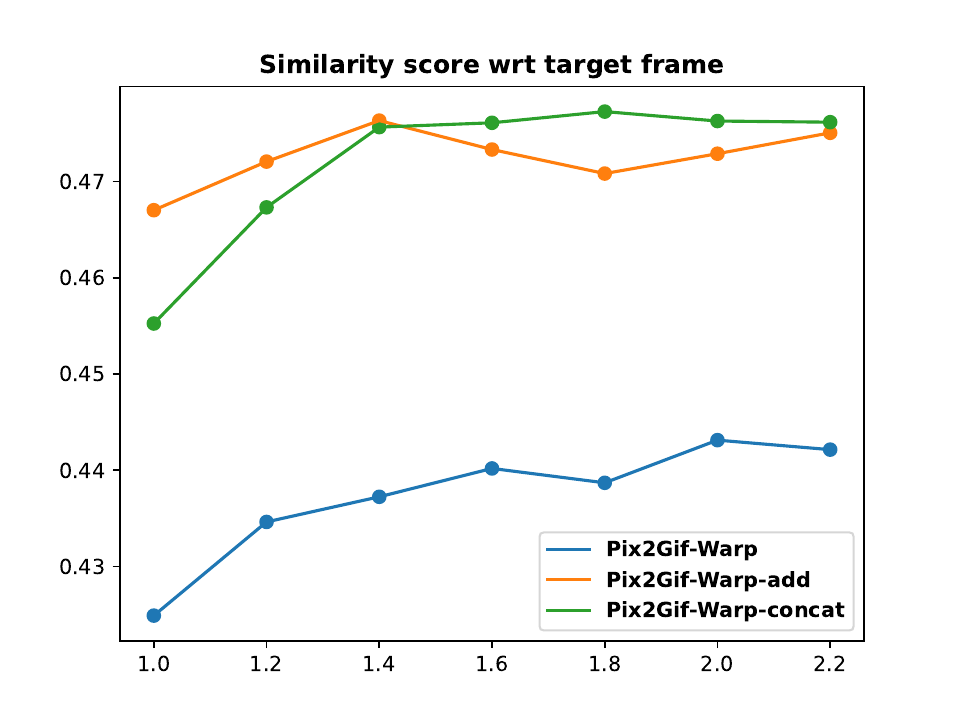}
  \caption{Target frame score}
  \label{fig:warp_tgt}
\end{subfigure}
\caption{Top: Ablation study between the earlier variants of our model by comparing average similarity score for 100 samples. Bottom: Ablation study on ways to input $z_W$ to $LDM$ by comparing average similarity score for 100 samples.}
\label{fig:abl_warp}
\end{figure*}

% \begin{figure*}
% \centering
% \begin{subfigure}{.25\textwidth}
%   \centering
%   \includegraphics[width=\linewidth]{images/src_abl.pdf}
%   \caption{Source frame score}
%   \label{fig:abl_src}
% \end{subfigure}%
% \begin{subfigure}{.25\textwidth}
%   \centering
%   \includegraphics[width=\linewidth]{images/tgt_abl.pdf}
%   \caption{Target frame score}
%   \label{fig:abl_tgt}
% \end{subfigure}
% \caption{Ablation study between the earlier variants of our model by comparing average similarity score for 100 samples.}
% \label{fig:abl_variants}
% \end{figure*}

% \begin{figure*}
% \centering
% \begin{subfigure}{.25\textwidth}
%   \centering
%   \includegraphics[width=\linewidth]{images/src_warp.pdf}
%   \caption{Source frame score}
%   \label{fig:warp_src}
% \end{subfigure}%
% \begin{subfigure}{.25\textwidth}
%   \centering
%   \includegraphics[width=\linewidth]{images/tgt_warp.pdf}
%   \caption{Target frame score}
%   \label{fig:warp_tgt}
% \end{subfigure}
% \caption{Ablation study on ways to input $z_W$ to $LDM$ by comparing average similarity score for 100 samples.}
% \label{fig:abl_warp}
% \end{figure*}

\subsection{Ablations} 

We design a few variants of \textit{Pix2Gif} for our ablation studies:
\begin{itemize}
    \item \textit{Pix2Gif-Base}: We train InstructPix2Pix with our data, and append the text prompt with ``The optical flow is \textunderscore.''.
    \item \textit{Pix2Gif-Motion-embed}: The motion embedding layer is added to encode the motion magnitude and combined with textual embedding.
    \item \textit{Pix2Gif-Warp}: We further add the warping module into the model but differently only use the warped feature for the $LDM$.
    \item \textit{Pix2Gif-Warp-add}: Different from \textit{Pix2Gif-Warp}, we instead add the warped feature and source image feature as input to the $LDM$.
    \item \textit{Pix2Gif-Warp-concat}: Instead of adding in \textit{Pix2Gif-Warp-add}, we concatenate the warped feature and source image feature as the input, but do not include the perceptual loss.
\end{itemize}

For comparative studies of our model with its variants, we generate an 8-frame video, with the motion input specified as [2, 4, 6, 8, 11, 14, 17, 19]. We extract features from our generated video using the X-CLIP model~\cite{ni2022expanding}, and employ CLIP to extract features from the source, target and generated frames. Throughout this discussion, we evaluate the performance of our model's variations using four metrics. We believe these metrics effectively measure the different facets of generating motion through the image translation framework utilized in \textit{Pix2Gif}.

\paragraph{Motion Coherency} 
Our task is formulated as an image translation problem, where we use motion magnitude as a guide. To evaluate the quality of the motion or temporal coherence in the GIFs, we calculate the L2 loss and Pearson Correlation Coefficient (PCC) for InstructPix2Pix, \textit{Pix2Gif-Base} and \ourmodel~.
The former evaluates how closely the motion values of the generated frames match the actual inputs, while the latter checks if they follow the same trend. These metrics are evaluated between the input motion magnitude values and the optical flow values, which are computed between the source image and the generated frames. As demonstrated in \cref{tab:coherency}, it is evident that \ourmodel~ has the highest correlation and the lowest L2 loss across all cfg\_img values, proving the effectiveness of our model architecture, which demonstrates its controllability in generating GIFs with specified motions. \textit{Pix2Gif-Base}, which uses the same architecture as InstructPix2Pix, shows much better performance than the original InstructPix2Pix. This highlights the significance of our new dataset.

\paragraph{Image-Video Similarity Score}
To evaluate the semantic characteristics of the video produced by different versions of our model, we devise two similarity scores: one in relation to the features of the source frame and another in relation to the features of the target frame.
 \begin{itemize}
     \item Source frame score: This score, in essence, quantifies the extent to which the generated video retains the fundamental attributes of the source frame. Thus, it measures the level of accuracy with which the source image is portrayed throughout the video sequence.
     \item Target frame score: This score serves as an indicator of the precision with which the generated video portrays the development of the scene or subject from the source frame. In addition, it highlights the model's capability of managing uncertainty and potential changes, as the target frame signifies a potential state that the generated video might attain.
 \end{itemize}
%The similarity score calculated between the video features and the source image features. This metric, in essence, quantifies the extent to which the generated video retains the fundamental attributes of the source image. Thus, it measures the level of accuracy with which the source image is portrayed throughout the video sequence.

%     \item Sim\_video\_target: The similarity score is calculated between the video features and the target image features. This metric serves as an indicator of the precision with which the generated video portrays the development of the scene or subject from the source frame. In addition, it highlights the model's capability of managing uncertainty and potential changes, as the target image signifies a potential state that the generated video might attain.

Our experimentation began with the \textit{Pix2Gif-Base}, followed by the integration of $\mathcal{M}$ which we denote as \textit{Pix2Gif-Motion-embed}, to improve the model's ability to condition on the motion value. We then added $\mathcal{W}$ to enable our model to learn from the optical flow feature map, and finally, we introduced $L_P$ to facilitate the generation of coherent latent features post-warping. These four models are compared in \cref{fig:abl_src} and \cref{fig:abl_tgt}. As can be seen, the \textit{Pix2Gif-Base} falls short when compared to the other iterations of our approach, as expected. In the optimal range of [1.6, 2.2], \textit{Pix2Gif} surpasses \textit{Pix2Gif-Warp-concat} and \textit{Pix2Gif-Motion-embed} because the motion generated by them is consistent with the input image but uncontrolled and with artifacts. While \textit{Pix2Gif} generates coherent and controlled motion, it often limits the extent of motion in comparison to the other two models because of $L_P$ which is a good trade off. As we cross the ideal range, we observe that all our models begin to converge, which is expected as the condition on input frame increases, the motion generated by the models decreases and they all begin to produce frames that are more or less identical. The \textit{Pix2Gif-Motion-embed} model creates a significant amount of one-directional motion, which can sometimes be nonsensical, and hence the addition of $\mathcal{W}$ helps to mitigate this issue.

In our discussion about the use of $\mathcal{W}$ to learn to translate features based on the source frame, we carried out a critical ablation study. This study focused on the effective utilization of the warped latent vector ($z_W$) for \textit{Pix2Gif} to function as it does currently. We experimented with three different \textit{Pix2Gif} variations to understand the optimal way of inputting $z_W$ into the $LDM$, with the quantitative comparison shown in \cref{fig:abl_warp}.   
In the first experiment, as illustrated in \cref{fig:model}, we fed only $z_W$ into the concat layer i.e. \textit{Pix2Gif-Warp}. However, this approach had a limitation as the $LDM$ lacked the original source image information and only possessed the shifted features from the source latent ($\mathcal{E}(c_{I})$). To address this, we experimented with including $\mathcal{E}(c_{I})$ in two ways: by adding $\mathcal{E}(c_{I})$ and $z_W$ before feeding them into the $LDM$ i.e. \textit{Pix2Gif-Warp-add}, or by concatenating $\mathcal{E}(c_{I})$ and $z_W$, which essentially represents \textit{Pix2Gif-Warp-concat}. As evident from \cref{fig:abl_warp}, the average similarity scores for both these models are higher than the \textit{Pix2Gif-Warp} version. Moreover, in the optimal range of [1.6, 2.2] for cfg$\_$img, \textit{Pix2Gif-Warp-concat} significantly outperforms \textit{Pix2Gif-Warp-add}. This can be attributed to the fact that in the addition process, $\mathcal{E}(c_{I})$ loses its unique characteristics, which are required by the diffusion model for effective unconditional denoising. Therefore, to achieve the best results, we combined $\mathcal{E}(c_{I})$ and $z_W$ before feeding them into the concat attention layer of the $LDM$.

% \begin{figure}[t]
%   \centering
%    \includegraphics[width=0.7\linewidth]{images/src_abl.pdf}

%    \caption{Ablation study between the earlier variants of our model by comparing similarity score between generated video and source image for 100 samples.}
%    \label{fig:abl_src}
% \end{figure}

% \begin{figure}[t]
%   \centering
%    \includegraphics[width=0.7\linewidth]{images/tgt_abl.pdf}

%    \caption{Ablation study between the earlier variants of our model by comparing similarity score between generated video and target image for 100 samples.}
%    \label{fig:abl_tgt}
% \end{figure}
\section{Conclusion}
\label{sec:conclusion}

In this work, we proposed \textit{Pix2Gif}, an image-to-GIF (video) generation model based on an image-to-image translation paradigm. To ensure temporal coherence across frames, we proposed a motion-guided warping module that learns to spatially warp the source image feature into the target one while maintaining visual consistency via a perceptual loss. Starting from TGIF, we curated a new dataset specifically used for training our model. The experimental results demonstrated the effectiveness of our model to generate GIFs with better temporal coherence compared with current state-of-the-art methods. Interestingly, the model also exhibits better controllability and some emerging action compositionality.

\section{Limitations and Future Work}
\label{sec:future_work}
The current \textit{Pix2Gif} model is our initial attempt to generate videos by treating it as an image translation task. However, this method has some limitations that prevent us from generating high-quality and long GIFs or videos. Firstly, the model generates images with a resolution of 256x256 pixels. If these images are used to generate subsequent frames, the quality of the frames deteriorates further. Secondly, due to limitations in computational power, we are only able to use a small portion of a larger, curated dataset for training our model. Our primary objective now is to improve the quality of the generated frames, as this could significantly enhance the effectiveness of this method.
{
    \small
    \bibliographystyle{ieeenat_fullname}
    \bibliography{main}
}


\end{document}